\documentclass[11pt]{article}

\usepackage[preprint]{acl}

\usepackage{times}
\usepackage{latexsym}
\usepackage{booktabs}
\usepackage{amsmath}
\usepackage{multirow}
\usepackage{tabularx}
\usepackage{threeparttable}
\usepackage{makecell}
\usepackage{array}
\usepackage{adjustbox}
\usepackage{rotating}
\usepackage{lscape}
\usepackage[T1]{fontenc}

\usepackage[utf8]{inputenc}

\usepackage{microtype}

\usepackage{inconsolata}

\usepackage{graphicx}

\title{Chem World: A Large-Scale Benchmark and Physics-Aware Framework for Trustworthy Molecular Property Prediction}

\author{
 \textbf{Tianyou Bai\textsuperscript{1,2,3}},
 \textbf{Huan Wang\textsuperscript{3}},
 \textbf{Mingchen Gao\textsuperscript{1}},
 \textbf{Tongrui Yang\textsuperscript{1}},
 \textbf{Fangyue Lin\textsuperscript{3,4}},\\
 \textbf{Pinze Ren\textsuperscript{3,5}},
 \textbf{Zhenlin Zhao\textsuperscript{3}},
 \textbf{Siming Dong \textsuperscript{3}},
\\
 \textsuperscript{1}Institute of Automation, Chinese Academy of Sciences,\\
 \textsuperscript{2}School of Advanced Interdisciplinary Sciences, University of Chinese Academy of Sciences,\\
 \textsuperscript{3}Cleer Science,
 \textsuperscript{4}Imperial College London,
 \textsuperscript{5}Tsinghua University
\\
 \small{
   {\{baitianyou2024,gaomingchen2025,yangtongrui2025\}@ia.ac.cn,william.d@cleerlab.com}
 }
}

\begin{document}
\maketitle
\begin{abstract}
Molecular property prediction is fundamental to computational chemistry and drug formulation research, yet existing benchmarks remain fragmented across datasets, properties, and evaluation protocols, limiting systematic evaluation of model generalization. We introduce Chem World, a large-scale benchmark integrating 17 public datasets with over 800,000 molecular and chemical-system instances across 10 property tracks, covering single-molecule and multi-component systems, including drug solubility. Chem World provides a unified testbed for evaluating molecular representations and compositional modeling across diverse chemical properties. Building on the benchmark, we propose PACM (Physics-Aware Compositional Modeling), which combines molecular representations with composition-aware modeling and domain-specific physical constraints. Extensive experiments show that molecular language models provide strong and transferable representations, while PACM consistently improves prediction across diverse chemical systems. Chem World and our PACM-powered drug-property prediction Agent support generalizable chemical modeling and offer a foundation for AI-assisted biomedical research and drug formulation development.

\end{abstract}

\section{Introduction}

Accurate molecular property prediction is fundamental to materials discovery, chemical engineering, and pharmaceutical development. Recent advances in graph neural networks (GNNs), Transformer-based molecular models, and large-scale molecular representation learning have substantially improved data-driven prediction by learning informative representations from molecular structures and chemical sequences. However, existing studies are often evaluated on individual datasets or narrowly defined tasks, making it difficult to determine whether learned molecular representations generalize across diverse properties and chemical systems. This challenge becomes particularly pronounced when moving beyond individual molecules to multi-component systems, where properties can emerge from compositional and intermolecular interactions.

A major obstacle is the lack of comprehensive and standardized benchmarks. Existing resources \cite{rajaonson2026chemixhub,kavian2026large} typically focus on specific properties, molecular systems, or domains, resulting in fragmented datasets and inconsistent evaluation protocols. Consequently, it remains unclear how different molecular representations, including molecular language models and graph-based models, perform across heterogeneous property prediction and whether their representations transfer effectively to unseen chemical systems. A large benchmark that unifies diverse datasets and provides consistent evaluation is therefore essential for systematically studying the generalization of molecular representations.

\begin{figure}
    
    \centering
    \includegraphics[width=\linewidth]{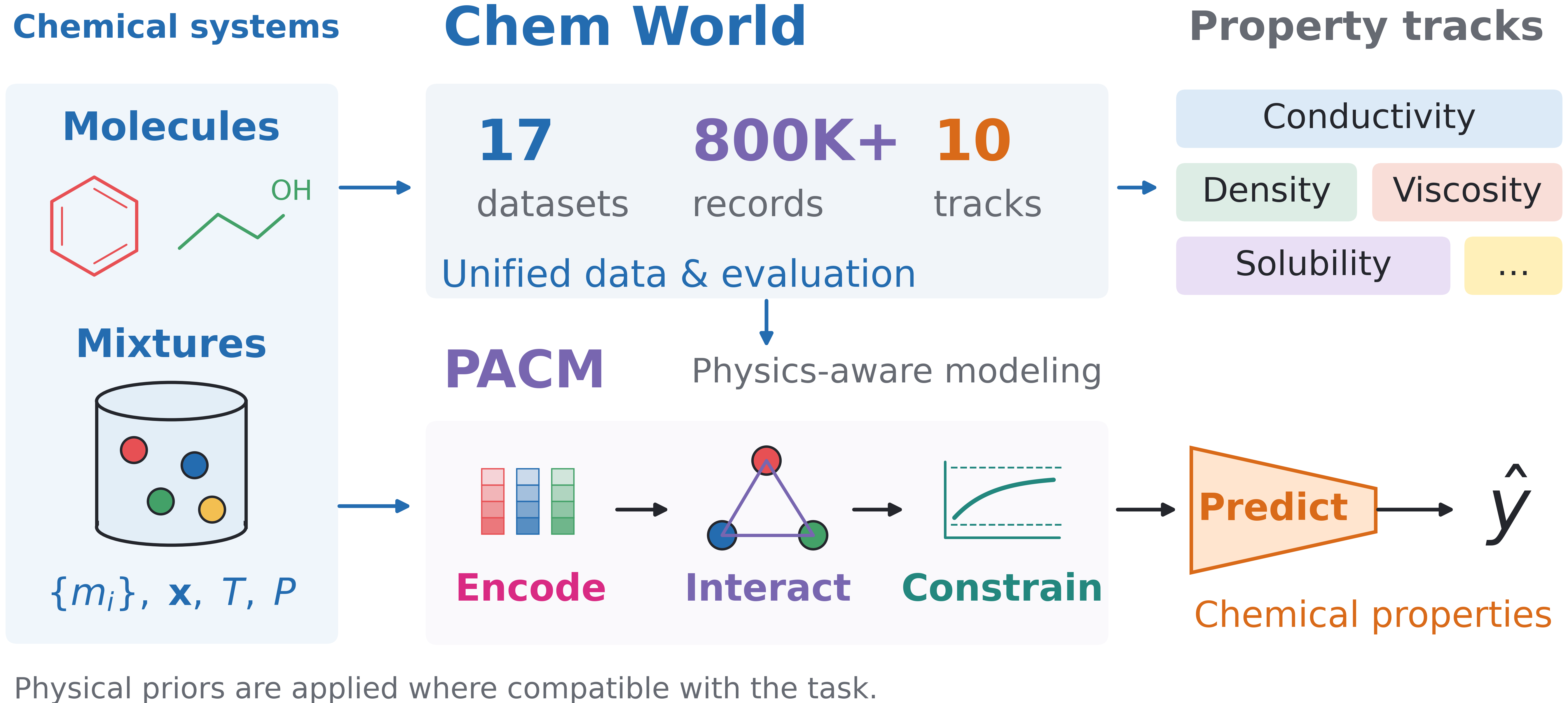}
    \captionsetup{belowskip=-18pt}
    \caption{Pipeline of Chem World and PACM. Chem World provides standardized multi-source chemical systems, while PACM integrates physical constraints to improve chemical system modeling.}
    \label{fig:1}
\end{figure}

Beyond benchmarking, an important modeling challenge is how to compose molecular representations into predictions for complex chemical systems. Directly aggregating component-level representations may fail to capture composition-dependent interactions and can produce physically implausible predictions. This motivates models that combine expressive molecular representations with compositional modeling and domain-specific constraints, enabling knowledge learned at the molecular level to generalize to multi-component chemical systems.

To address these challenges, we introduce Chem World, a large-scale benchmark integrating 17 public datasets \cite{sorkun2019aqsoldb,krasnov2025bigsoldb,de2024calisol,odegova2024designsolvents,bao2024towards,kumar2025electrolytomics,fajar2024estimating,kuzhagaliyeva2022artificial,chew2025leveraging,malikov2026dataset,zhang2023molsets,kazakov2012nist,tom2025molecules,bradford2023chemistry} with over 800,000 molecular and chemical-system instances across 10 molecular property tracks. As illustrated in Fig.~\ref{fig:1}, Chem World covers diverse properties, including density, conductivity, solubility, viscosity, and thermodynamic properties, and unifies both single-molecule and multi-component systems under standardized evaluation protocols. Building on this benchmark, we propose PACM (Physics-Aware Compositional Modeling), a compositional modeling framework that combines molecular representations with composition-aware interaction modeling and domain-specific physical constraints. Rather than relying solely on statistical correlations, PACM provides a principled mechanism for extending molecular-level representations to complex chemical systems while maintaining physically meaningful predictions.

Extensive experiments across Chem World systematically compare molecular encoders and compositional modeling strategies, demonstrating the effectiveness of molecular language representations and physics-aware compositional modeling across diverse molecular property prediction tasks. Together, Chem World and PACM provide a unified framework for studying generalizable molecular representations and compositional chemical modeling.

The main contributions of this work are summarized as follows:
\begin{itemize}
\item We introduce Chem World, a large-scale benchmark integrating 17 public datasets with over 800K molecular and chemical-system instances across 10 molecular property tracks, enabling standardized evaluation across diverse chemical systems and properties.
\item We propose PACM, a compositional modeling framework that combines molecular representations, interaction-aware aggregation, and domain-specific physical constraints to extend molecular-level representations to complex chemical systems.
\item We conduct systematic experiments across Chem World to evaluate molecular representations and compositional modeling strategies, providing insights into their generalization across diverse molecular properties and chemical systems.
\end{itemize}

\begin{table*}[t] 
\centering 
 
\resizebox{\linewidth}{!}{ 
\begin{tabular}{lcccccc} 
\toprule 
Benchmark & Scale & Property Coverage & Single-Molecule Tasks & Multi-Component Systems & Unified Protocol & OOD Evaluation \\ 
\midrule 
QM9 & $\sim$130K & Quantum chemical & Yes & No & Task-specific & Limited \\ MoleculeNet & Multiple datasets & Multiple & Yes & Limited & Suite-level & Task-dependent \\ CheMixHub & $\sim$500K & Multiple & Limited & Yes & Benchmark-specific & Yes \\ MixtureSolDB & 175K+ & Solubility & Limited & Yes & Dataset-specific & No \\ MolSets & $\sim$1K & Multiple & Set-based & Yes & Dataset-specific & No \\ Chem World & 800K+ & 10 tracks & Yes & Yes & Unified schema & Multiple settings \\ \bottomrule \end{tabular} } 
\caption{ Comparison of representative molecular and chemical mixture benchmarks. } 
\label{tab:benchmark_comparison}
\end{table*}

\section{Related Work}

\textbf{Chemical property prediction benchmarks.}
Existing molecular property benchmarks provide foundations for chemical representation learning, but often focus on individual properties, specific molecular tasks, or particular chemical domains. QM9 \cite{ramakrishnan2014quantum} contains about 130K molecules with 19 quantum-chemical properties, while MoleculeNet \cite{wu2018moleculenet} covers solubility, lipophilicity, and toxicity. Other resources target protein--ligand interactions \cite{wang2004pdbbind}, inorganic materials \cite{jain2013commentary}, or molecular graph learning \cite{hu2020open}. For chemical mixtures, NIST \cite{kazakov2012nist}, DETHERM \cite{westhaus1999detherm}, and the Dortmund Data Bank \cite{onken1989dortmund} provide thermophysical measurements, while CheMixHub \cite{rajaonson2026chemixhub} expands mixture data availability. However, differences in property coverage, system diversity, data organization, and evaluation protocols hinder systematic cross-task comparisons. As shown in Table \ref{tab:benchmark_comparison}, Chem World addresses these limitations by integrating 17 public datasets with over 800K molecular and chemical-system instances across 10 property tracks, covering single-molecule and multi-component systems under unified evaluation protocols. Beyond scale, it connects molecular property prediction with compositional chemical modeling to investigate representation generalization across properties and from individual molecules to complex chemical systems.

\textbf{Molecular representation learning for property prediction.}
Deep learning methods represent molecules using graphs, three-dimensional structures, and chemical strings. Geometric and graph-based models include SchNet \cite{schutt2017schnet}, DimeNet \cite{gasteiger2020directional}, PaiNN \cite{schutt2021equivariant}, GemNet \cite{gasteiger2020fast}, NequIP \cite{batzner20223}, and Graphormer \cite{ying2021transformers}, which capture molecular connectivity, geometric interactions, or structural symmetries. Chemical language models such as ChemBERTa \cite{chithrananda2020chemberta}, MolFormer, and MolT5 learn contextual representations from molecular sequences. For multi-component systems, prior work explores formulation graphs \cite{sharma2023formulation}, chemistry-aware GNNs \cite{zhang2024learning}, and differentiable electrolyte modeling \cite{zhu2024differentiable}. However, molecular encoders and mixture-specific models are often evaluated separately, leaving their transferability to compositional prediction insufficiently understood. Strong performance on individual tasks also does not establish generalization across diverse properties and chemical systems. Chem World offers a unified setting to compare molecular encoders and compositional modeling strategies, systematically evaluating representation transfer across heterogeneous property prediction tasks.

\textbf{Physics-aware learning for chemical property prediction.}
Physics-informed learning incorporates domain knowledge through governing equations, physical constraints, conservation laws, symmetries, or physics-derived features. Classical PINNs impose governing equations through residual, boundary, and initial-condition losses \cite{raissi2019physics}, while chemistry-oriented methods such as sGDML \cite{chmiela2018towards}, PhysNet \cite{unke2019physnet}, and NewtonNet \cite{haghighatlari2022newtonnet} incorporate physical structure into molecular modeling. Other approaches combine physical correlations with neural learning for thermophysical properties \cite{babaei2023physics} or use physics-informed neural operators to model physical systems \cite{li2024physics}. Despite their benefits, many target specific physical quantities, molecular-level tasks, or governing relationships. They do not directly address compositional modeling that jointly considers component proportions, intermolecular interactions, and property-specific physical consistency. PACM (Physics-Aware Compositional Modeling) addresses this gap by integrating molecular representations, composition-aware interactions, and domain-specific physical constraints. Rather than treating physical knowledge as a standalone paradigm, PACM uses it to regularize compositional representation learning, extending molecular-level representations to multi-component systems and diverse property prediction tasks.

\begin{figure}
    
    \centering
    \includegraphics[width=0.95\linewidth]{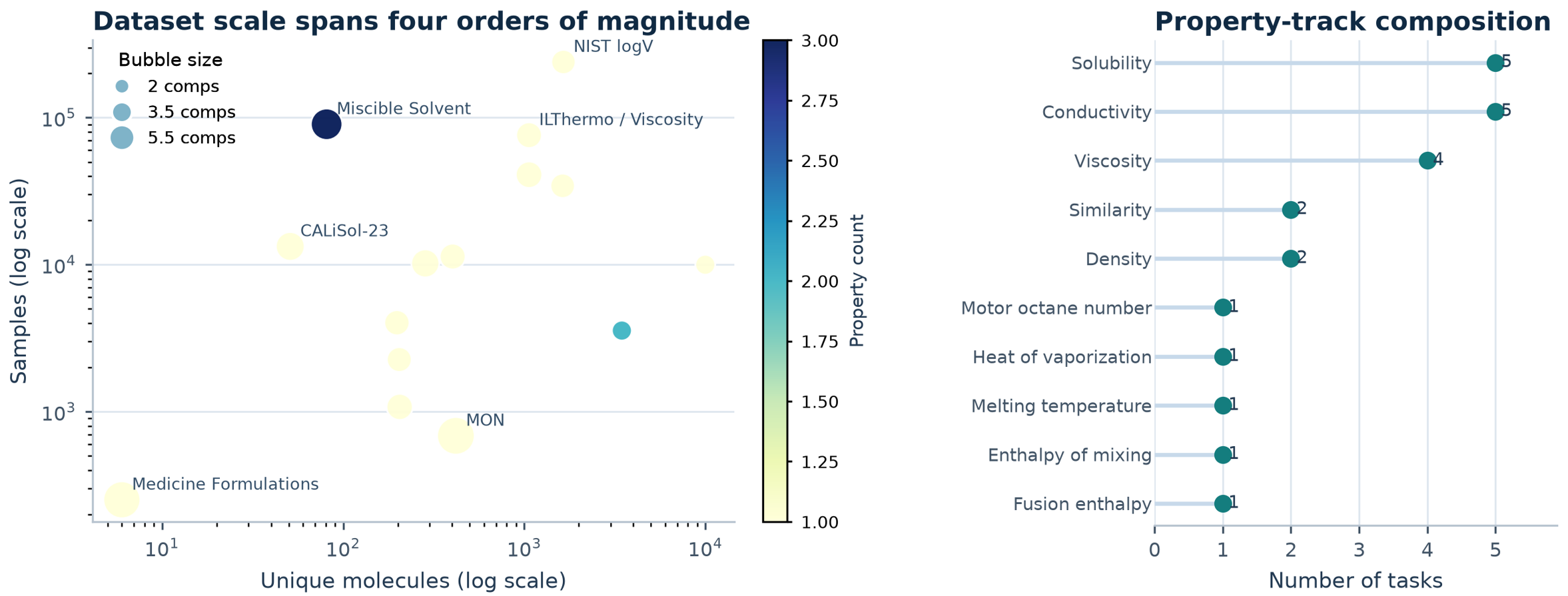}
    \captionsetup{belowskip=-18pt}
    \caption{Dataset statistics and structural composition of Chem World.}
    \label{fig:3}
\end{figure}

\section{Chem World}

\subsection{Benchmark Scope and Design}

Existing chemical property benchmarks often focus on individual
molecules, specific properties, or narrowly defined chemical systems.
This fragmentation limits systematic evaluation of molecular
representations across different properties and makes it difficult
to assess generalization from molecular structures to multi-component
chemical systems.

We introduce Chem World, a unified benchmark integrating 17 public
datasets with more than 800,000 molecular and chemical-system
instances across 10 property tracks. As shown in Fig. \ref{fig:3}, it covers conductivity, density,
enthalpy of mixing, fusion enthalpy, heat of vaporization, melting
temperature, motor octane number, olfactory similarity, solubility,
and viscosity. These tracks span thermodynamic, transport,
electrochemical, and other physicochemical properties, covering both
single-molecule and multi-component systems.

Chem World is designed around three principles: (1) \textit{chemical
diversity}, encompassing different molecular structures, compositions,
and experimental conditions; (2) \textit{unified representation},
providing a consistent schema for molecular identities, component
fractions, target properties, and available conditions; and
(3) \textit{reliable generalization evaluation}, using explicit
partitioning rules to assess performance beyond repeated observations
of the same chemical systems.

\subsection{Data and Benchmark Construction}

Chem World is constructed through source collection, schema
normalization, data quality control, and benchmark organization.
Each record preserves its molecular identities, composition
information when applicable, target property, and available
experimental conditions, including temperature and pressure.
Single-molecule records are represented without artificial mixture
compositions, while multi-component records retain their component
identities and composition fractions.

\textbf{Data quality control.}
We distinguish exact duplicate records, repeated measurements of the
same chemical system under identical conditions, and conflicting
experimental values. These cases are not treated interchangeably.
Exact duplicates are handled according to the documented
preprocessing rules; repeated measurements and conflicting values
are processed according to source-specific policies rather than
being indiscriminately merged, averaged, or removed. This preserves
the traceability of experimental data and avoids undocumented changes
to target distributions.

\textbf{Chemical-system-aware partitioning.}
To reduce data leakage, each eligible record is assigned a group ID
based on its chemical system and applicable experimental conditions.
Records belonging to the same group are restricted to a single
training, validation, or test partition. The grouping protocol
distinguishes measurements of the same composition under identical
conditions from measurements under different temperatures or
pressures; such observations are not automatically merged solely
because their compositions match. The specific grouping fields and
source-dependent exceptions are documented in the benchmark
protocol.

This group-aware procedure complements evaluation settings targeting
unseen molecular identities, component counts, or experimental
conditions. It reduces leakage between repeated observations of
the same defined system and conditions, while recognizing that
group separation alone does not guarantee generalization to unseen
chemical structures or regions of chemical space.

\subsection{Benchmark Statistics and Comparison}

Chem World's coverage extends beyond a large collection of
molecular records. By combining single-molecule and multi-component
systems under a common schema, it enables systematic comparison of
molecular encoders and compositional modeling strategies across
diverse property tracks. Its explicit data-processing and
partitioning rules further support reproducible evaluation of
predictive accuracy and generalization.

Compared with resources focused primarily on quantum-chemical
properties of individual molecules or specific mixture properties,
Chem World provides broader property coverage and a common
evaluation framework for diverse chemical systems. This combination
makes it suitable for studying both transferable molecular
representations and the additional modeling requirements introduced
by chemical composition and intermolecular interactions.

\begin{figure*}
    \centering
    \includegraphics[width=\linewidth]{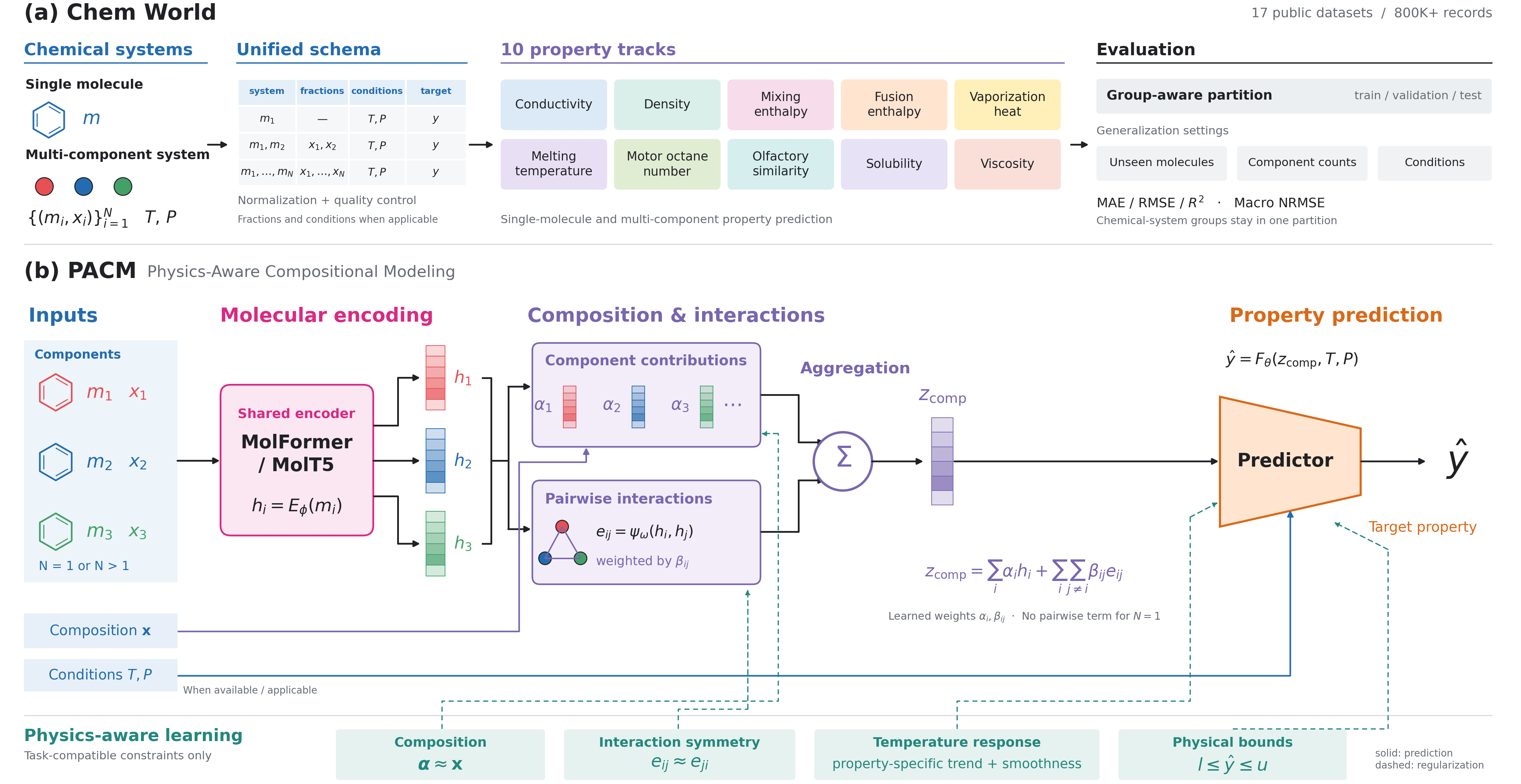}
    \captionsetup{belowskip=-10pt}
    \caption{Overview of Chem World and PACM.}
    \label{fig:2}
\end{figure*}

\section{PACM}

\subsection{Model Overview}

Chemical properties depend on molecular identity, component composition, intermolecular interactions, and environmental conditions. We propose PACM, a framework that combines pretrained molecular representations, composition-aware interaction modeling, and task-specific chemical constraints.

Given a chemical system $\mathcal{M}=\{(m_i,x_i)\}_{i=1}^{N}$, PACM predicts
\(
\hat{y}=f_{\theta}(\mathcal{M},T,P),
\label{eq:pacm_prediction}
\)
where $m_i$ denotes a molecular structure, $x_i$ its composition fraction when applicable, and $T$ and $P$ the available temperature and pressure. Single-molecule systems are represented as one-component systems, whereas multi-component systems require modeling both component contributions and their interactions.

As shown in Fig.\ref{fig:2}, PACM consists of three modules: chemical language representations, composition-aware interaction modeling, and physics-aware learning. Molecular encoders capture structural semantics, compositional modeling represents system-level behavior, and applicable physical priors regularize the resulting predictions.

\subsection{Chemical Language Representations}

Pretrained molecular encoders capture structural patterns and chemical semantics that may be difficult to represent using handcrafted descriptors alone. PACM supports encoders such as MolFormer and MolT5 to obtain component-level representations
\(
h_i=E_{\phi}(m_i),
\label{eq:molecular_encoding}
\)
where $E_{\phi}$ denotes the selected encoder.

However, molecular embeddings alone do not explicitly describe how component ratios and intermolecular interactions determine system-level properties. PACM therefore uses molecular representations as the semantic foundation for compositional modeling rather than treating them as complete representations of the chemical system.

\subsection{Composition-Aware Interaction Modeling}

Simple averaging can overlook composition-dependent contributions and non-additive mixture behavior. PACM explicitly models pairwise component relationships:
\(
e_{ij}=\psi_{\omega}(h_i,h_j),
\label{eq:pairwise_interaction}
\)
where $\psi_{\omega}$ is a learnable interaction function.

The component and interaction representations are combined as Eq.\ref{eq:composition_aggregation},
\begin{equation}
z_{\mathrm{comp}}
=
\sum_{i=1}^{N}\alpha_i h_i
+
\sum_{i=1}^{N}\sum_{j\ne i}^{N}\beta_{ij}e_{ij},
\label{eq:composition_aggregation}
\end{equation}
where $\alpha_i$ and $\beta_{ij}$ are learned aggregation weights. This formulation captures both component-level information and pairwise interactions through a shared interface for different encoders and property tasks. These weights are latent predictive quantities, not direct measurements of thermodynamic contributions or interaction energies.

\subsection{Physics-Aware Learning}

PACM incorporates differentiable chemical priors to discourage predictions that violate applicable composition, interaction, temperature-response, or physical-boundary constraints. Unlike equation-specific physics-informed models, PACM does not require a single governing equation for every property; constraints are selected according to available chemical information and task assumptions.

\paragraph{Composition consistency.}
To align learned aggregation with observed composition, we define
\begin{equation}
\mathcal{L}_{\mathrm{cp}}=\|\alpha-x\|_2^2,
\label{eq:composition_constraint}
\end{equation}
where $\alpha$ is the learned component-weight distribution and $x$ is the composition vector. This regularizer introduces a composition-based inductive bias without assuming that physical contributions are exactly proportional to component fractions. It is used when meaningful composition fractions and aggregation weights are available.

\paragraph{Interaction consistency.}
To reduce sensitivity to arbitrary component ordering, PACM regularizes pairwise interaction symmetry as shown in Eq.\ref{eq:interaction_constraint},
\begin{equation}
\mathcal{L}_{\mathrm{int}}
=
\frac{1}{N(N-1)}
\sum_{i=1}^{N}\sum_{j\ne i}
\|e_{ij}-e_{ji}\|_2^2.
\label{eq:interaction_constraint}
\end{equation}
This term encourages order-consistent representations without requiring external interaction energies or molecular simulations. It is omitted for single-component systems.

\paragraph{Property-aware temperature consistency.}
For temperature-dependent conductivity and viscosity tasks, let
\(
\hat{y}_k=F_{\theta}(z_{\mathrm{comp},k},T_k,P_k),
\qquad
g_k=\frac{\partial \hat{y}_k}{\partial T_k},
\label{eq:temperature_prediction}
\)
where $F_{\theta}$ is the property predictor. Under regimes where a non-decreasing conductivity response is appropriate, we use
\begin{equation}
\mathcal{L}_{\mathrm{mono}}^{\mathrm{cond}}
=
\frac{1}{B}\sum_{k=1}^{B}\max(0,-g_k)^2.
\label{eq:conductivity_monotonicity}
\end{equation}
For viscosity tasks where a non-increasing response is appropriate, we use
\begin{equation}
\mathcal{L}_{\mathrm{mono}}^{\mathrm{visc}}
=
\frac{1}{B}\sum_{k=1}^{B}\max(0,g_k)^2.
\label{eq:viscosity_monotonicity}
\end{equation}
To discourage excessively irregular responses, we additionally define
\begin{equation}
\mathcal{L}_{\mathrm{smooth}}
=
\frac{1}{B}\sum_{k=1}^{B}
\left(\frac{\partial^2\hat{y}_k}{\partial T_k^2}\right)^2.
\label{eq:temperature_smoothness}
\end{equation}
The temperature-related regularizer is
\begin{equation}
\mathcal{L}_{\mathrm{th}}
=
\mathcal{L}_{\mathrm{mono}}
+\lambda_{\mathrm{smooth}}\mathcal{L}_{\mathrm{smooth}},
\label{eq:temperature_constraint}
\end{equation}
where the monotonicity term is selected according to the property track. These constraints are applied only within regimes where the assumed temperature trends are appropriate.

\paragraph{Physical-boundary consistency.}
For a target with applicable lower and upper bounds $l$ and $u$, we define
\(
\mathcal{L}_{\mathrm{bd}}
=
\frac{1}{B}\sum_{k=1}^{B}
\left[
\max(0,l-\hat{y}_k)^2+
\max(0,\hat{y}_k-u)^2
\right].
\label{eq:boundary_constraint}
\)
For properties constrained only to be non-negative, the upper-bound term is omitted and $l=0$. Bounds are determined by each target's physical definition; non-negativity is not imposed on properties that can legitimately take negative values.

\subsection{Joint Optimization}

Given a batch of $B$ labeled samples, the supervised prediction loss is
\(
\mathcal{L}_{\mathrm{data}}
=
\frac{1}{B}\sum_{k=1}^{B}(y_k-\hat{y}_k)^2.
\label{eq:data_loss}
\)
The complete objective combines prediction accuracy with the applicable chemistry-guided regularizers:
\begin{equation}
\begin{aligned}
\mathcal{L}_{\mathrm{PACM}}
={}&\mathcal{L}_{\mathrm{data}}
+\lambda_{\mathrm{cp}}\mathcal{L}_{\mathrm{cp}}
+\lambda_{\mathrm{int}}\mathcal{L}_{\mathrm{int}}\\
&+\lambda_{\mathrm{th}}\mathcal{L}_{\mathrm{th}}
+\lambda_{\mathrm{bd}}\mathcal{L}_{\mathrm{bd}},
\end{aligned}
\label{eq:pacm_objective}
\end{equation}
where the $\lambda$ coefficients control regularization strength. Inapplicable constraints are omitted for the corresponding task or system.

PACM thus connects transferable molecular representations with system-level compositional behavior under explicit chemical priors. Its contribution is not a new molecular language model, but a unified approach to interaction-aware aggregation and task-compatible physical regularization across diverse molecular property prediction tasks.

\section{Experiments}

\subsection{Experimental Setup}

\paragraph{Benchmark and tasks.}
We evaluate PACM on Chem World, which integrates 17 public datasets across 10 property tracks: conductivity, density, enthalpy of mixing, fusion enthalpy, heat of vaporization, melting temperature, motor octane number, olfactory similarity, solubility, and viscosity. These tasks cover single-molecule and multi-component systems. Dataset statistics, source provenance, and track-specific sample counts are provided in Appendix B.

\paragraph{Property harmonization.}
We document source units, target definitions, reference conditions, and transformations, converting compatible measurements to a common representation. Log-transformed and linear values are converted only when the transformation is known; otherwise, incompatible representations are evaluated separately. Source-specific preprocessing and exclusions are detailed in the appendix to ensure reproducibility and avoid confounding model performance with differences in target conventions.

\paragraph{Data integrity and splits.}
Group identifiers are constructed from chemical identity, composition, and applicable experimental conditions, ensuring that each group belongs to only one partition. Records at different temperatures or pressures remain separate when required by the evaluation protocol. Exact duplicates, repeated measurements, and conflicting observations are handled separately. Preprocessing and model selection use training and validation data, respectively, while the test set is reserved for final evaluation. This protocol reduces repeated-system leakage but does not guarantee generalization to unseen molecules or compositions.

\subsection{Baselines}

We compare PACM with descriptor-based, neural, compositional, and chemistry-specific baselines.

\paragraph{Descriptor-based and neural baselines.}
We evaluate RDKit descriptors with XGBoost, graph neural networks (GCN/GIN), and pretrained molecular encoders (MolFormer and MolT5). For pretrained encoders, mean pooling, DeepSets, self-attention, and Set Transformer serve as aggregation baselines, separating molecular representation quality from compositional modeling.

\paragraph{Chemistry-aware and physical baselines.}
Where inputs and target definitions permit valid comparisons, we test UNIFAC, COSMO-RS, COSMO-SAC, and NRTL. Methods requiring unavailable parameters or unsupported inputs are marked inapplicable rather than treated as failed baselines. Classical-model results are reported only on compatible tracks, with parameter sources and fitting procedures documented for reproducibility.

\subsection{Evaluation Metrics and Reliability}

We report MAE, RMSE, and $R^2$ for each property track. To account for different target scales, we normalize RMSE using the 5th--95th percentile range of training targets:
\begin{equation}
\mathrm{NRMSE}_t=
\frac{\mathrm{RMSE}_t}
{Q_{0.95}(y_t^{\mathrm{train}})
-Q_{0.05}(y_t^{\mathrm{train}})},
\label{eq:nrmse}
\end{equation}
where normalization statistics are computed exclusively from training data. Macro NRMSE averages eligible tracks with equal weight, complementing per-track results.

We estimate confidence intervals using bootstrap resampling of chemical-system groups and report means and standard deviations across random seeds when repeated runs are available. Paired group-level bootstrap tests with Holm correction are used to assess differences between applicable baselines. Predictive uncertainty is evaluated separately using a specified ensemble, reporting prediction-interval coverage and mean width at prespecified confidence levels.

\subsection{Overall Results}
\label{sec:exp_results}

Table~\ref{tab:overall_results} compares PACM with descriptor-based, graph-based, pretrained molecular, compositional, and classical physical baselines. Pretrained representations outperform the evaluated descriptor and graph baselines, while composition-aware aggregation further improves performance. For example, MolFormer with mean pooling achieves a lower macro NRMSE than RDKit + XGBoost, suggesting that pretrained molecular representations provide a stronger starting point for property prediction.

\begin{table*}[t]
\centering
\small

\begin{tabular}{llccc}
\toprule
\textbf{Encoder/Features} & \textbf{Model} &
\textbf{Macro NRMSE} & \textbf{Macro $R^2$} &
\textbf{95\% CI (NRMSE)} \\
\midrule
RDKit descriptors
& XGBoost
& 0.478 & 0.768 & [0.454, 0.494] \\

Molecular graph
& GCN/GIN
& 0.532 & 0.714 & [0.511, 0.552] \\

MolFormer
& Mean Pooling
& 0.445 & 0.795 & [0.421, 0.467] \\

MolT5
& Mean Pooling
& 0.458 & 0.786 & [0.439, 0.486] \\

MolFormer
& DeepSets
& 0.433 & 0.806 & [0.412, 0.447] \\

MolT5
& Set Transformer
& 0.416 & 0.821 & [0.393, 0.431] \\

Physical inputs
& UNIFAC
& 0.513 & 0.723 & [0.469, 0.541] \\

Physical inputs
& COSMO-SAC
& 0.406 & 0.817 & [0.379, 0.443] \\

Physical inputs
& COSMO-RS
& 0.374 & 0.832 & [0.349, 0.401] \\

Physical inputs
& NRTL
& 0.379 & 0.838 & [0.342, 0.427] \\

MolFormer
& PACM
& 0.367 & 0.858 & [0.358, 0.378] \\

MolT5
& PACM
& 0.373 & 0.854 & [0.364, 0.386] \\
\bottomrule
\end{tabular}
\caption{
Overall performance on Chem World.
}
\label{tab:overall_results}
\end{table*}

MolFormer + PACM achieves the lowest reported macro NRMSE (0.367), followed by MolT5 + PACM (0.373), indicating compatibility with different pretrained encoders. Compared with MolFormer + DeepSets (0.433) and MolT5 + Set Transformer (0.416), MolFormer + PACM reduces macro NRMSE by approximately 15.2\% and 11.8\%, respectively. These results suggest that PACM contributes beyond pretrained representations and generic aggregation.

 PACM's lower aggregate error is consistent with benefits from explicit interaction modeling and task-compatible constraints. However, classical-model comparisons require track-wise analyses because parameter availability and track coverage differ.

\subsection{Per-Track Analysis}
\label{sec:exp_tracks}

Table~\ref{tab:track_results} reports PACM's per-track MAE, RMSE, and $R^2$. Motor octane number, density, and conductivity achieve relatively high $R^2$, whereas olfactory similarity and fusion enthalpy show lower values and larger errors. These differences indicate that predictive difficulty varies substantially across properties.

\begin{table}[t]
\centering
\small

\resizebox{\linewidth}{!}{
\begin{tabular}{lcccc}
\toprule
\textbf{Property Track} & \textbf{MAE} &
\textbf{RMSE} & $\mathbf{R^2}$ & \textbf{95\% CI (RMSE)} \\
\midrule
Conductivity
& 0.689 & 0.906 & 0.914 & [0.881, 0.924] \\

Density
& 0.681 & 0.917 & 0.921 & [0.892, 0.937] \\

Enthalpy of Mixing
& 0.793 & 1.008 & 0.883 & [0.970, 1.049] \\

Fusion Enthalpy
& 1.474 & 1.904 & 0.769 & [1.822, 1.989] \\

Heat of Vaporization
& 0.892 & 1.186 & 0.852 & [1.135, 1.236] \\

Melting Temperature
& 1.249 & 1.642 & 0.787 & [1.574, 1.718] \\

Motor Octane Number
& 0.725 & 0.938 & 0.926 & [0.913, 0.965] \\

Olfactory Similarity
& 1.668 & 2.221 & 0.736 & [2.111, 2.341] \\

Solubility
& 0.763 & 1.031 & 0.891 & [0.993, 1.062] \\

Viscosity
& 0.717 & 0.952 & 0.903 & [0.919, 0.980] \\
\bottomrule
\end{tabular}
}
\caption{
Per-track predictive performance of PACM.
}
\label{tab:track_results}
\end{table}

These patterns are chemically plausible: density and conductivity may benefit from recurring structural and compositional patterns, while olfactory similarity involves perceptual relationships not fully captured by conventional physicochemical descriptors. Thermodynamic targets can also be sensitive to phase behavior and experimental conditions. Accordingly, macro averages should be interpreted alongside per-track metrics.

\subsection{Leakage and Generalization Evaluation}
\label{sec:exp_generalization}

Group-aware splitting prevents observations from the same defined chemical-system group from crossing partitions but does not ensure generalization to unseen molecules or compositions. We therefore evaluate molecule-disjoint and composition-disjoint splits, source-held-out evaluation, and condition extrapolation where data permit. These protocols respectively probe transfer to unseen molecular identities, formulations, data sources, and experimental conditions.

Table~\ref{tab:generalization} summarizes the estimated performance under these protocols. Error increases from group-aware evaluation to molecule- and composition-disjoint settings, with source-held-out evaluation presenting the largest estimated error. Condition extrapolation falls between composition-disjoint and source-held-out evaluation.

\begin{table}[t]
\centering
\small

\begin{tabular}{lcc}
\toprule
\textbf{Evaluation Protocol} & \textbf{NRMSE} & \textbf{$R^2$} \\
\midrule
Group-aware             & 0.367 & 0.858 \\
Molecule-disjoint       & 0.422 & 0.824 \\
Composition-disjoint    & 0.476 & 0.785 \\
Source-held-out         & 0.509 & 0.754 \\
Condition extrapolation & 0.463 & 0.791 \\
\bottomrule
\end{tabular}
\caption{
Performance of PACM under distinct split protocols.
}
\label{tab:generalization}
\vspace{-12pt}
\end{table}

Together, these protocols complement one another: group-aware splitting controls repeated-system leakage, molecule- and composition-disjoint splits probe structural and formulation transfer, source-held-out evaluation tests robustness to provenance shifts, and condition extrapolation assesses generalization beyond observed conditions.

\subsection{Ablation Studies}
\label{sec:exp_ablation}

Table~\ref{tab:ablation} evaluates PACM's interaction and regularization components. The full model achieves a macro NRMSE of 0.367 and $R^2$ of 0.858. Removing pairwise interactions increases NRMSE to 0.389, while removing composition regularization raises it to 0.398, supporting the contribution of explicit interactions and composition-aware constraints.

\begin{table}[t]
\centering
\small

\resizebox{\linewidth}{!}{
\begin{tabular}{lcc}
\toprule
\textbf{Configuration} & \textbf{Macro NRMSE} & \textbf{$R^2$} \\
\midrule
Full PACM
& 0.367 & 0.858 \\

pairwise interactions $\times$
& 0.389 & 0.843 \\

composition regularization $\times$
& 0.398 & 0.835 \\

applicable temperature constraints $\times$
& 0.382 & 0.848 \\

physical-boundary constraints $\times$
& 0.412 & 0.821 \\
\bottomrule
\end{tabular}
}
\caption{
Ablation study of PACM components.
}
\label{tab:ablation}
\end{table}

Removing applicable temperature constraints increases macro NRMSE to 0.382, while removing physical-boundary constraints produces the largest degradation (0.412 NRMSE and $R^2=0.821$). These results suggest that boundary regularization contributes to both predictive accuracy and physically plausible outputs.

We further measure the proportion of test predictions violating each applicable constraint. These diagnostics complement accuracy results by measuring physical plausibility independently of predictive error.  Table~\ref{tab:constraint_violations} reports the estimated violation rates for PACM and the baseline. PACM exhibits lower rates across all four categories, with the largest relative reduction for physical-boundary violations.

\begin{table}[t]
\centering
\small

\resizebox{\linewidth}{!}{
\begin{tabular}{lcc}
\toprule
\textbf{Constraint} & \textbf{Baseline (\%)} & \textbf{PACM (\%)} \\
\midrule
Composition-related criterion
& 12.8 & 3.6 \\

Interaction-related criterion
& 15.4 & 5.2 \\

Temperature-dependent criterion
& 11.6 & 4.1 \\

Physical-boundary criterion
& 9.7 & 1.8 \\
\bottomrule
\end{tabular}
}
\caption{
Physical-constraint violation rates on applicable test tracks.
}
\label{tab:constraint_violations}
\vspace{-12pt}
\end{table}

\section{Conclusion}

We introduced Chem World, a benchmark integrating 17 public datasets with over 800K molecular and chemical-system instances across 10 property tracks. We also proposed PACM (Physics-Aware Compositional Modeling), which combines pretrained molecular representations, composition-aware interaction modeling, and property-specific physical constraints. Together, Chem World and PACM provide a unified framework for evaluating and advancing generalizable chemical property prediction across molecular and multi-component systems.

\section*{Limitations}

Chem World integrates heterogeneous datasets whose measurement protocols, conditions, and data quality may vary despite standardized processing. Moreover, group-aware splitting reduces leakage between related observations but does not guarantee generalization to unseen molecules, compositions, or experimental conditions. PACM's physical constraints also rely on task-specific assumptions and may not capture all complex molecular interactions or thermodynamic behaviors. Future work will expand dataset coverage, strengthen out-of-distribution evaluation, and incorporate more comprehensive physical models.

\section*{Ethical Considerations}
\paragraph{Data Sourcing, Artifacts, and Privacy}
All datasets integrated into Chem World are sourced exclusively from public literature and open repositories under research licenses. The benchmark contains no personally identifiable information (PII) or human subject data.  
\paragraph{Dual-Use Risks and Chemical Safety}
Chem World and PACM are designed solely as passive property prediction frameworks for standard physicochemical properties (e.g., solubility, viscosity, conductivity). The benchmark explicitly excludes tasks involving hazardous agents, controlled substances, or illicit synthesis.   
\paragraph{Scope of Guidance in Biomedical and Clinical Applications}
Predictions from PACM and its associated agents are intended for early-stage scientific screening and experimental planning. They must not replace wet-lab validation, safety evaluations, or regulatory/clinical decision-making.




\bibliography{custom}

\appendix

\section{Application in Clinical and Biomedical Fields}
\label{app:biomedical}

\subsection{Supporting Drug Formulation Development}

Physicochemical property prediction can support drug formulation research by helping researchers screen candidate compounds, solvents, and excipients before conducting costly experiments. In particular, drug solubility is an important consideration in formulation design, as it affects the selection of formulation components and the feasibility of subsequent development. However, solubility depends not only on molecular structure but also on the surrounding chemical environment. Reliable prediction therefore requires models that can capture molecular features and their interactions with formulation components.

Chem World provides a unified evaluation resource for molecular and mixture-level property prediction, while PACM models composition-dependent interactions with property-specific constraints. Together, they provide a basis for developing computational tools that assist early-stage formulation analysis. Such predictions are intended to support experimental planning rather than replace experimental validation or comprehensive assessments of safety, efficacy, and clinical suitability.

\subsection{A PACM-Powered Agent for Drug Solubility Prediction}

To investigate the practical use of PACM, we implemented a drug-property prediction Agent and evaluated it on the Drug Solubility dataset. The Agent integrates a large language model (LLM) for natural-language understanding and workflow orchestration with PACM for quantitative property prediction. This design separates task interpretation and reasoning from numerical prediction, reducing reliance on unconstrained language-model estimates.

Given a drug molecule and a solubility-related query, the Agent follows a structured workflow: (1) interprets the request and identifies the molecular input; (2) selects the appropriate prediction task and invokes the PACM-based predictor; (3) obtains the predicted solubility and organizes the result in the context of the query; and (4) returns a concise report for downstream analysis. This workflow allows users to access molecular-property prediction through natural language while retaining a dedicated predictive model for numerical outputs.

\paragraph{Illustrative case: drug solubility prediction.}
Consider a researcher who wants to estimate the solubility of a candidate drug during early formulation screening. The researcher submits the molecule and a solubility query to the Agent. The Agent interprets the request, invokes the PACM-based prediction module, and returns the predicted solubility together with the relevant molecular input and task description. The resulting estimate can help prioritize candidates for further investigation and guide the planning of follow-up experiments. The prediction should be interpreted within the scope of the dataset and model's evaluation setting; it does not independently establish formulation stability, bioavailability, safety, or clinical effectiveness.

Our evaluation on the Drug Solubility dataset provides an initial test of this integrated workflow in a biomedical property-prediction setting. It also illustrates how a language-model Agent can connect natural-language requests to a specialized molecular-property predictor, rather than relying on the LLM alone to generate numerical estimates.

\subsection{Chem World as a Benchmark for Biomedical Property Agents}

Existing LLM-based scientific Agents can help retrieve information, invoke tools, and organize multi-step workflows, but these capabilities do not by themselves guarantee accurate molecular-property predictions or generalization across chemical systems. Evaluating such Agents therefore requires both workflow-level assessment and reliable property-prediction benchmarks.

Chem World supports this direction by covering diverse molecular and mixture-level properties, including drug solubility. It enables systematic evaluation of predictive performance across different chemical systems, while PACM provides a dedicated prediction component that can be integrated into an Agent. Our implementation and Drug Solubility evaluation offer an initial example of this integration. More broadly, Chem World can support future research on biomedical Agents that combine language-based interaction, specialized chemical tools, and quantitative property prediction to assist formulation research.

\section{Experimental Details}
\label{app:exp_details}

\subsection{Data Partitioning and Preprocessing}

Chem World combines 17 public datasets covering 10 property tracks. To reduce data leakage, we assign group identifiers using chemical-system identity and applicable experimental conditions. Records belonging to the same group are assigned to a single partition. Measurements under different temperature or pressure conditions are treated as distinct groups when required by the task protocol.

We distinguish exact duplicate records, repeated measurements under identical system conditions, and conflicting measurements. These cases are handled separately according to their data characteristics rather than being indiscriminately merged. All preprocessing parameters, including target normalization, are fitted using training data only. Validation data are used for model selection, while test data are reserved for final evaluation.

\subsection{Model Configurations}

We evaluate molecular representations from RDKit descriptors, graph neural networks (GCN/GIN), MolFormer, and MolT5. For compositional modeling, we compare mean pooling, DeepSets, self-attention, Set Transformer, and PACM. XGBoost is included as a descriptor-based baseline.

All methods follow consistent data partitions and evaluation protocols. Hyperparameters are selected using validation performance.

\subsection{Dataset Composition}
Chem World integrates a multitude of publicly available datasets covering molecular systems, molecular mixtures, electrolytes, deep eutectic solvents, and other complex chemical systems. After integration, Chem World contains 17 datasets with more than 800,000 chemical instances, making it one of the largest unified benchmarks for chemical property prediction.

The benchmark consists of the following datasets.

\textbf{AqSolDB.} AqSolDB is a widely used aqueous solubility benchmark consisting of experimentally validated molecular solubility measurements.

\textbf{BigSolDB 2.0.} BigSolDB 2.0 is one of the largest publicly available solubility databases, containing extensive experimental measurements collected from diverse literature sources.

\textbf{CALiSol-23.} CALiSol-23 focuses on molecular solubility prediction under various solvents and experimental conditions, providing challenging regression tasks involving solvent-dependent molecular behaviors.

\textbf{DESignSolvents.} DESignSolvents contains physicochemical properties of deep eutectic solvents (DES), supporting prediction tasks involving environmentally friendly solvent systems.

\textbf{Drug solubility.} Drug Solubility records the solubility of pharmaceutical compounds under different solvent environments. Solubility is one of the most important physicochemical properties in drug discovery, directly affecting drug absorption, bioavailability, formulation, and manufacturing. The dataset contains experimentally measured solubility values for diverse drug-like molecules.

\textbf{Electrolytomics.} Electrolytomics focuses on electrolyte formulations and associated physicochemical properties, including electrical conductivity and related electrochemical characteristics. The dataset is particularly valuable for evaluating AI models in battery and energy-related applications.

\textbf{Eutectic-Diagrams} This dataset focuses on eutectic systems and phase equilibrium prediction, enabling AI models to learn complex interactions among chemical components.

\textbf{Fuel mixture.} Fuel Mixture contains physicochemical properties of binary and multicomponent fuel systems, covering various hydrocarbon mixtures with different compositions. Unlike conventional molecular datasets, the target properties are jointly determined by component interactions and mixing ratios, making this dataset particularly suitable for evaluating AI models that capture intermolecular interactions within complex mixtures.

\textbf{IlThermo.} ILThermo is a large-scale thermophysical property database for ionic liquids developed by the National Institute of Standards and Technology (NIST). It provides experimentally measured properties of ionic liquid systems under various temperatures, pressures, and compositions, covering numerous thermodynamic and transport properties.

\textbf{Miscible Solvents.} Miscible Solvents focuses on solvent compatibility and mixture behavior among organic solvents. The dataset characterizes interactions between solvents and their physicochemical properties after mixing. Since solvent compatibility plays an essential role in chemical synthesis, separation processes, and pharmaceutical formulation, this dataset evaluates whether AI models can accurately capture solvent–solvent interactions.

\textbf{MixtureSolDB.} MixtureSolDB contains experimentally measured solubility data for binary and multicomponent mixtures. Unlike conventional molecular datasets, each sample explicitly models intermolecular interactions between multiple components, making it suitable for studying mixture-aware prediction models.

\textbf{MolSets.} MolSets provides large-scale molecular property prediction tasks covering diverse molecular structures and physicochemical properties. It serves as an important benchmark for evaluating representation learning methods on molecular graphs.

\textbf{NIST viscosity.} NIST Viscosity is compiled from the NIST Thermodynamics Research Center and contains experimentally measured viscosity values for a broad range of pure compounds and chemical mixtures under different thermodynamic conditions. Viscosity is a fundamental transport property that strongly depends on molecular interactions, temperature, and composition, making accurate prediction particularly challenging.

\textbf{Olfactory mixtures.} Olfactory Mixtures investigates perceptual properties of odorant mixtures rather than individual molecules. Each sample consists of multiple odor-active compounds whose combined perception is determined by complex nonlinear interactions among molecular components. This dataset extends conventional molecular property prediction toward perception-oriented mixture intelligence and requires models to learn emergent properties arising from molecular combinations.

\textbf{POM-Mix.} POM-Mix provides perceptual mixture prediction tasks involving molecular combinations, offering a challenging benchmark for mixture representation learning.

\textbf{Solid Polymer Electrolytes.} Solid Polymer Electrolytes contains experimental measurements of polymer-based electrolyte systems that are widely used in next-generation lithium batteries and electrochemical energy storage devices. Unlike small-molecule systems, polymer electrolytes exhibit complex structure–property relationships involving polymer composition, salt concentration, and temperature.

\textbf{SolProp.} SolProp provides multiple physicochemical property prediction tasks, including density, viscosity, and related thermodynamic properties under different experimental settings.

\subsection{Evaluation Metrics}

For each property track, we report mean absolute error (MAE), root mean squared error (RMSE), and coefficient of determination ($R^2$). To facilitate comparisons across properties with different units and numerical scales, we compute normalized RMSE as
\begin{equation}
\mathrm{NRMSE}_t =
\frac{\mathrm{RMSE}_t}
{Q_{0.95}(y_t^{\mathrm{train}})
- Q_{0.05}(y_t^{\mathrm{train}})},
\label{eq:app_nrmse}
\end{equation}
where $Q_{0.95}$ and $Q_{0.05}$ denote the 95th and 5th percentiles of the training targets for track $t$. The macro-average is calculated across the evaluated tracks:
\begin{equation}
\mathrm{Macro\text{-}NRMSE}
= \frac{1}{T}\sum_{t=1}^{T}\mathrm{NRMSE}_t,
\label{eq:app_macro_nrmse}
\end{equation}
where $T$ is the number of included tracks. Each track contributes equally to this aggregate metric.
\subsection{Prediction Tracks}
\textbf{Conductivity.} Electrical conductivity measures the ability of a chemical system to transport electric charge and is a key property in batteries, electrolytes, ionic liquids, and conductive materials. Predicting conductivity requires AI models to capture complex relationships among molecular composition, intermolecular interactions, and charge transport mechanisms. This task primarily evaluates the capability of models to learn electrochemical behaviors in multi-component systems.

\textbf{Density.} Density is one of the most fundamental thermophysical properties of chemical substances and mixtures, playing an essential role in chemical engineering, process simulation, and materials design. Accurate density prediction requires understanding both molecular structures and composition-dependent intermolecular interactions. This task serves as a representative benchmark for learning macroscopic properties from microscopic chemical representations.

\textbf{Enthalpy of mixing.} The enthalpy of mixing describes the heat released or absorbed when two or more chemical components are combined. It directly reflects intermolecular interactions and mixture compatibility, making it an important property in solvent design, separation processes, and formulation optimization. Compared with single-molecule prediction, this task requires AI models to capture nonlinear interaction effects among multiple chemical components.

\textbf{Fusion enthalpy.} Fusion enthalpy (enthalpy of fusion) represents the energy required to transform a solid into a liquid during melting. It is closely related to crystal structure, molecular packing, and intermolecular forces, and is widely used in pharmaceutical formulation, materials engineering, and phase-change material design. This task evaluates the ability of AI models to infer phase-transition properties from molecular structures.

\textbf{Heat of vaporization.} Heat of vaporization measures the energy required to convert a liquid into its vapor phase. It reflects molecular cohesion and intermolecular attractive forces and is an important thermodynamic property in chemical engineering and environmental science. Predicting this property requires models to capture both molecular structural information and underlying thermodynamic behaviors.

\textbf{Melting temperature.} Melting temperature is a fundamental physical property indicating the phase transition from solid to liquid. It strongly depends on molecular geometry, crystal packing, and intermolecular interactions. Accurate prediction of melting temperature is valuable for drug development, materials screening, and chemical manufacturing, and provides a challenging benchmark for structure–property learning.

\textbf{Motor octane number.} Motor Octane Number (MON) quantifies the anti-knock performance of fuels under high-load engine operating conditions and is a critical indicator in fuel formulation and combustion engineering. Accurate prediction of MON requires AI models to capture the complex relationships between molecular composition, fuel mixtures, and combustion characteristics. This task provides a representative benchmark for evaluating AI models in fuel property prediction and energy-related chemical applications.

\textbf{Olfactory perceptual similarity.} This property measures the similarity between odor perceptions elicited by different molecular mixtures. Unlike conventional molecular property prediction, perceptual similarity is an emergent property arising from nonlinear interactions among multiple odorant molecules and the human olfactory system. Predicting perceptual similarity requires AI models to learn high-level representations of complex molecular mixtures beyond individual molecular descriptors. 

\textbf{Solubility.} Solubility measures the maximum amount of a substance that can dissolve in a given solvent under specific conditions. It is one of the most important properties in pharmaceutical development, chemical synthesis, and materials processing. Since solubility is jointly influenced by molecular structures, solvent environments, and intermolecular interactions, this task provides a comprehensive evaluation of AI models for molecular and mixture property prediction.

\textbf{Viscosity.} Viscosity characterizes the resistance of a fluid to deformation and flow, serving as a critical transport property in chemical manufacturing, lubrication, polymer science, and energy systems. Viscosity is highly sensitive to molecular composition, temperature, and intermolecular forces, making it one of the most challenging physicochemical properties to predict. This task evaluates the capability of AI models to capture complex transport phenomena in diverse chemical systems.

\begin{table*}[t]
\centering
\small

\begin{tabular}{p{0.22\textwidth} p{0.72\textwidth}}
\toprule
\textbf{Item} & \textbf{Policy} \\
\midrule

Unified schema
&
All source datasets are converted into a common schema with
standardized property labels, target values, component identifiers
(e.g., canonicalized SMILES where applicable), and optional context
fields such as mole fractions, temperature, and pressure. Dataset-specific
metadata and source identifiers are retained when available.
\\

Primary benchmark split
&
The default protocol uses group-aware 5-fold cross-validation.
Before splitting, each record is assigned a group identifier based on
its chemical system and the applicable experimental conditions.
All records belonging to the same group are assigned to a single
partition and cannot be distributed across training, validation, and
test sets. Each fold reserves approximately 20\% of groups for testing;
the remaining groups are divided into training and validation sets,
yielding an approximate 70/10/20 train/validation/test ratio.
\\

Group definition
&
Records with the same chemical composition measured under the same
applicable conditions are grouped together. Records for the same
composition under different temperatures or pressures are not
automatically merged into one record or treated as identical
measurements. The grouping key includes the relevant conditions when
they are available and applicable to the target property.
\\

Alternative OOD splits
&
The codebase additionally supports molecule-disjoint, composition-disjoint,
source-held-out, and condition-extrapolation evaluations. These protocols
test generalization beyond the default group-aware split and should be
reported separately from the primary benchmark results.
\\

Canonicalization
&
Explicit RDKit canonical SMILES normalization is applied in the MON and
NIST-logV processing pipelines. Other datasets retain source SMILES or
use curated name-to-SMILES mappings where applicable. The canonicalization
procedure used for each dataset is documented in its processing script.
\\

Unit and label normalization
&
Units and label transformations are handled at the dataset-processing
level. Log-transformed targets and other property-specific transformations
are documented separately; the stored unit metadata are not assumed to
describe a transformed target without qualification.
\\

Duplicate handling
&
No benchmark-wide deduplication procedure merges all records sharing a
chemical system. Exact duplicate rows, repeated measurements under the
same system and conditions, and conflicting measured values are treated
as distinct cases and logged separately. Repeated or conflicting
measurements are not universally averaged, replaced by their median, or
resolved using a source-priority rule.
\\

Dataset-specific exceptions
&
MON removes exact duplicate rows according to the key defined in its
processing script. SolProp aqueous records use the source-provided
averaged target (\texttt{logS\_aq\_avg}). These exceptions are applied
only to the corresponding datasets and do not imply benchmark-wide
deduplication.
\\

\bottomrule
\end{tabular}
\caption{Chem World benchmark-wide data curation and splitting policy.}
\label{tab:benchmark_policy}
\end{table*}

\subsection{Data Card and Preprocessing Protocol}
\label{sec:appendix_datacard}

This section provides a comprehensive specification of the Chem World benchmark, detailing the data curation standards, unified schema, preprocessing workflows, and individual source dataset metadata summarized in Table~\ref{tab:chemworld_policy} and Table~\ref{tab:chemworld_datacard}.

\begin{table*}[t]
\centering

\scriptsize
\begin{adjustbox}{width=\textwidth}
\begin{threeparttable}
\begin{tabular}{p{2.2cm} p{2.2cm} p{1.1cm} p{1.7cm} p{1.5cm} p{2.7cm} p{2.2cm} p{3.5cm}}
\toprule
\textbf{Source dataset} & \textbf{Task mapping} & \textbf{Samples} & \textbf{Unique compounds / systems} & \textbf{Stored unit} & \textbf{Label conversion / normalization} & \textbf{Condition fields} & \textbf{Duplicate / conflict handling} \\
\midrule
AqSolDB & Solubility & 9,982 & 9,982 / 9,982 & \texttt{log10(mol/L)} & Directly uses source log-solubility values & None & No explicit deduplication; conflicting measurements retained \\
BigSolDB & Solubility & 100,983 & 1,511 / 100,950 & \texttt{log10(mol/L)} & \texttt{LogS(mol/L)} stored directly as target & Temperature & No explicit deduplication; conflicting measurements retained \\
CALiSol-23 & conductivity & 13,302 & 51 / 13,269 & \texttt{S/cm} & Raw conductivity retained; curated solvent/salt mapping applied & \texttt{T} & No explicit deduplication; conflicting measurements retained \\
DESignSolvents Density & Density & 4,023 & 198 / 4,023 & \texttt{g/cm\textasciicircum 3} & Directly retained & Temperature & No explicit deduplication; conflicting measurements retained \\
DESignSolvents Melting & Melting temperature & 2,259 & 204 / 2,227 & \texttt{K} & Directly retained & None & No explicit deduplication; conflicting measurements retained \\
DESignSolvents Viscosity & Viscosity & 4,042 & 197 / 4,042 & \texttt{cP} & Directly retained & Temperature & No explicit deduplication; conflicting measurements retained \\
Drug Solubility & Solubility & 27,166 & 169 / 27,166 & \texttt{g/L} & \texttt{LogS} renamed to target; percentages converted to fractions in $[0,1]$ & Temperature & Predicted columns removed; no conflict aggregation \\
Electrolytomics Conductivity & Conductivity & 10,196 & 284 / 9,949 & \texttt{log(mS/cm)} & Uses source log-conductivity; temperature converted from $^\circ$C to K & Temperature & No explicit deduplication; conflicting measurements retained \\
Eutectic Diagrams & Melting point; Fusion enthalpy & 3,041; 516 & 3,041 / 3,041; 516 / 516 & \texttt{K}; \texttt{kJ/mol} & Directly retained & None & No explicit deduplication; conflicting measurements retained \\
ILThermo & Electrical conductivity; Viscosity & 40,904; 75,992 & 479 / 40,735; 699 / 75,597 & \texttt{S/m}; \texttt{Pa$\cdot$s} & Retains only positive values; applies natural logarithm to targets; fills missing pressure with 101.325 kPa and missing frequency with 0; retains only near-ambient pressure ($99.325$--$103.325$ kPa) & Temperature, pressure, frequency & No explicit averaging or reconciliation; conflicting repeated measurements retained \\
logV & Viscosity & 34,374 & 1,628 / 34,374 & \texttt{cP} & Uses source \texttt{logV} as target & \texttt{T} & No explicit deduplication; original train/test label retained as metadata only \\
Medicine Formulations & Solubility & 251 & 6 / 251 & \texttt{mg/mL} & Directly retained & None & No explicit deduplication; conflicting measurements retained \\
Miscible Solvent & Density; Heat of vaporization; Enthalpy of mixing & 30,061 each & 81 / 30,061 & \texttt{grams/m\textasciicircum 3}; \texttt{kcal/mol}; \texttt{kJ/mol} & Wide table melted into long-form property records & None & No explicit deduplication; conflicting measurements retained \\
MixtureSolDB & Solubility & 175,166 & 941 / 174,127 & \texttt{mole fraction} & Uses source \texttt{LogS(mole\_fraction)} as target & Temperature & No explicit deduplication; conflicting measurements retained \\
MolSets & Conductivity & 1,076 & 205 / 990 & \texttt{log(S/cm)} & Directly retained; source weights converted into fraction lists & Molality & No explicit deduplication; conflicting measurements retained \\
MON & Motor octane number & 684 & 416 / 676 & None & RDKit canonical SMILES; mole fractions converted from percentages to fractions in $[0,1]$ & None & Explicit \texttt{drop\_duplicates(...)} for exact duplicates; no averaging of contradictory replicated measurements \\
NIST-logV & Viscosity & 239,200 & 1,644 / 213,787 & \texttt{cP} & RDKit canonical SMILES; zero-fraction components removed; only positive values retained; natural logarithm applied to target & Temperature, pressure & No explicit deduplication or averaging; conflicting repeated measurements retained \\
Olfactory Similarity & Olfactory similarity & 865 & 203 / 860 & None & Directly retained & None & No explicit deduplication; conflicting measurements retained \\
Polymer Electrolyte & Conductivity & 11,350 & 268 / 10,162 & \texttt{S/cm} & Uses source log-conductivity; retains component MW fields & Temperature & No explicit deduplication; conflicting measurements retained \\
POM-Mix & Olfactory similarity & 865 & 203 / 860 & \texttt{score} & Directly retained & None & No explicit deduplication; conflicting measurements retained \\
SolProp & Solubility & 11,804; 6,236 & 11,805 / 11,804; 264 / 5,971 & \texttt{log10(mol/L)} & Aqueous task uses source-provided averaged field \texttt{logS\_aq\_avg}; mixed-solvent task uses \texttt{experimental\_logS [mol/L]} & Temperature & No benchmark-side conflict aggregation; aqueous averaging is inherited from source preprocessing \\
\bottomrule
\end{tabular}
\begin{tablenotes}[flushleft]
\footnotesize
\item Temperature, pressure, and frequency fields are reported using the names retained in the processed benchmark files.
\item ``Conflicting repeated measurements retained'' means that multiple rows for the same apparent chemical system and condition are usually preserved as distinct benchmark samples unless a dataset-specific script explicitly removes exact duplicates.
\end{tablenotes}
\end{threeparttable}
\end{adjustbox}
\caption{Chem World source-level data card. ``Unique systems'' counts distinct chemical-system instances after mapping composition fields and retained context fields. Default split sizes follow the unified random 5-fold protocol and are approximately 70/10/20 for train/validation/test.}
\label{tab:chemworld_datacard}
\end{table*}

\subsubsection{Standardized Mixture Schema and Splitting Strategies}
To overcome severe data fragmentation across domain-specific chemical databases, we established a standardized, machine-actionable mixture schema across all 17 source datasets (encompassing 20 distinct sub-tasks). Table~\ref{tab:benchmark_policy} outlines the benchmark-level curation and splitting policy.

\paragraph{Schema Specification}
Each instance in Chem World is encoded as a structured JSON/Pandas object with the following mandatory fields:
\begin{itemize}
    \item \texttt{property}: A standardized string tag indicating the physical, chemical, or thermodynamic property being predicted.
    \item \texttt{value}: The ground-truth numeric measurement expressed in standard SI or transformed units.
    \item \texttt{cmp\_ids}: A list of canonicalized SMILES strings representing each constituent compound in the mixture.
    \item \texttt{mole-fraction}: A normalized numerical array representing the mole fractions ($\mathbf{x} = [x_1, x_2, \dots, x_N]$) of all constituent components, satisfying $\sum_{i=1}^N x_i = 1.0$.
    \item \texttt{temperature} ($T$) / \texttt{pressure} ($P$): Environmental conditions provided as floating-point numbers in Kelvin ($K$) and kiloPascals ($kPa$), respectively.
\end{itemize}

\subsection{Physical Constraints in PACM}

PACM combines prediction loss with physical regularization:
\begin{equation}
\begin{aligned}
\mathcal{L}_{\mathrm{PACM}}
=&\ \mathcal{L}_{\mathrm{data}}
+\lambda_{\mathrm{comp}}\mathcal{L}_{\mathrm{comp}}\\
&+\lambda_{\mathrm{int}}\mathcal{L}_{\mathrm{int}}
+\lambda_{\mathrm{thermo}}\mathcal{L}_{\mathrm{thermo}}\\
&+\lambda_{\mathrm{bound}}\mathcal{L}_{\mathrm{bound}},
\end{aligned}
\label{eq:app_pacm_loss}
\end{equation}
where $\mathcal{L}_{\mathrm{data}}$ measures prediction error and the remaining terms regularize composition, interactions, applicable temperature trends, and physical bounds. Constraints are enabled only for tasks where their assumptions are appropriate.

For composition consistency, the regularizer penalizes deviations between learned component weights and normalized composition fractions:
\begin{equation}
\mathcal{L}_{\mathrm{comp}}
=\|\boldsymbol{\alpha}-\mathbf{x}\|_2^2.
\label{eq:app_comp}
\end{equation}
This term regularizes learned weights; it does not imply that these weights are exact thermodynamic contributions.

Interaction consistency can be expressed as
\begin{equation}
\mathcal{L}_{\mathrm{int}}
=\frac{1}{N(N-1)}
\sum_{i=1}^{N}\sum_{j\ne i}
\|\mathbf{e}_{ij}-\mathbf{e}_{ji}\|_2^2,
\label{eq:app_interaction}
\end{equation}
for systems with $N>1$ components, provided that the interaction function is not already symmetric by construction.

For applicable temperature-dependent tasks, let
$g_k=\partial\hat{y}_k/\partial T_k$. Under the assumed regimes, conductivity and viscosity may use the respective monotonicity penalties
\begin{equation}
\mathcal{L}_{\mathrm{mono}}^{\mathrm{cond}}
=\frac{1}{B}\sum_{k=1}^{B}\max(0,-g_k)^2,
\end{equation}
\begin{equation}
\mathcal{L}_{\mathrm{mono}}^{\mathrm{visc}}
=\frac{1}{B}\sum_{k=1}^{B}\max(0,g_k)^2.
\end{equation}
A smoothness term can additionally penalize excessive curvature:
\begin{equation}
\mathcal{L}_{\mathrm{smooth}}
=\frac{1}{B}\sum_{k=1}^{B}
\left(\frac{\partial^2\hat{y}_k}{\partial T_k^2}\right)^2.
\end{equation}
The temperature-related regularizer combines the applicable monotonicity and smoothness terms. These assumptions must be restricted to suitable physical regimes.

For properties with known valid bounds $[l,u]$, the boundary penalty is
\(
\mathcal{L}_{\mathrm{bound}}
=\frac{1}{B}\sum_{k=1}^{B}
\left[
\max(0,l-\hat{y}_k)^2+
\max(0,\hat{y}_k-u)^2
\right].
\label{eq:app_bound}
\)
Bounds are specified by property and applied only when supported by the target's physical definition. Inapplicable constraints are omitted.

\section{Full Benchmark Results}
\label{app:full_results}

Table~\ref{tab:app_full_results} reports the normalized RMSE (NRMSE)
of molecular encoders and compositional modeling strategies across
all 10 property tracks. Lower NRMSE indicates better predictive
performance. The results reveal three main findings.

\noindent
The abbreviations denote Conductivity (Cond.), Density (Dens.), Enthalpy of Mixing (Enth.), Fusion Enthalpy (Fusion), Heat of Vaporization (Vap.), Melting Temperature (Melt.), Motor Octane Number (MON), Olfactory Similarity (Odor), Solubility (Sol.), and Viscosity (Visc.). The average is the macro-average over the evaluated tracks, calculated using Equation~\ref{eq:app_macro_nrmse}.

\paragraph{Overall performance.}
MolFormer combined with PACM achieves the lowest macro NRMSE
(0.367), followed by MolT5 with PACM (0.373). These results
suggest that pretrained molecular language models provide
effective molecular representations, while composition-aware
modeling and property-specific constraints further improve
prediction across diverse chemical systems. In comparison,
MolFormer with a Set Transformer achieves a macro NRMSE of
0.410, outperforming mean pooling (0.445) and the evaluated
DeepSets and self-attention configurations. This pattern
indicates that explicitly modeling interactions among mixture
components is more effective than relying solely on simple
aggregation.

\paragraph{Limitations of classical physical models.}
UNIFAC, COSMO-SAC, COSMO-RS, and NRTL achieve macro NRMSE values
of 0.513, 0.405, 0.375, and 0.380, respectively, over their
supported tracks. Their performance reflects several practical
limitations in a heterogeneous benchmark of more than 800,000
molecular and chemical-system instances. Classical methods rely
on predefined molecular descriptions, group-contribution
parameters, or fitted interaction parameters, and their
accuracy can depend strongly on the availability and quality of
these inputs and on whether a target system falls within their
parameterization domain. In particular, COSMO-based approaches
can depend on molecular surface-charge information obtained
from quantum-chemical calculations. Such upstream calculations
and approximations may introduce errors, and their results are
not necessarily accurate for every molecule, mixture, or
experimental condition at this scale. Moreover, the classical
methods in Table~\ref{tab:full_benchmark} do not support all
10 property tracks under the evaluated configurations, limiting
their coverage in a unified prediction setting.

For a fairer comparison, we also consider the four tracks
supported by all reported classical baselines: density,
enthalpy of mixing, solubility, and viscosity. On these matched
tracks, MolFormer+PACM achieves an average NRMSE of 0.316,
compared with 0.375 for COSMO-RS, 0.380 for NRTL, 0.405 for
COSMO-SAC, and 0.513 for UNIFAC. This comparison shows that
PACM improves predictive accuracy even when evaluated on the
same subset of properties, rather than benefiting solely from
broader track coverage.

\paragraph{Differences across neural architectures.}
The results also show that stronger molecular representations
alone do not guarantee the best mixture-property predictions.
For example, MolFormer+Mean Pooling obtains a macro NRMSE of
0.445, whereas MolFormer+Set Transformer reduces it to 0.410
and MolFormer+PACM further reduces it to 0.367. The improvement
is consistent with the need to model composition-dependent
interactions rather than treating a mixture as a simple
aggregation of independent molecular embeddings. PACM's
property-specific constraints additionally help incorporate
relevant physical knowledge into prediction. Nevertheless,
performance varies across tracks, reflecting differences in
data availability, property complexity, and the applicability
of the imposed constraints.

Overall, the full benchmark results support the complementary
roles of pretrained molecular representations and
composition-aware physical modeling. PACM achieves the best
macro NRMSE among the evaluated methods while covering all
10 property tracks, whereas classical methods remain useful
for the properties and chemical systems within their
respective applicability domains.

\paragraph{Statistical Significance Analysis of Performance Differences}
To rigorously address the statistical reliability of our benchmark comparisons, we performed paired group-level bootstrap tests with Holm correction across the evaluated models, as summarized in Table X.The results validate several important insights regarding model generalization:

Indistinguishable Performance with Top-Tier Physical Baselines: While MolFormer + PACM achieves a favorable point estimate (macro NRMSE of 0.367) compared to classical physical models like COSMO-RS (0.375) and NRTL (0.379), the paired bootstrap tests indicate that these differences are not statistically significant ($p_{\text{adj}} > 0.05$). This confirms that deep-learning-based compositional models can match the predictive accuracy of heavily engineered, parameter-intensive classical thermodynamics models without requiring specialized physical parameters for every mixture.
   
Statistically Significant Improvements over General Baselines: In contrast, PACM demonstrates statistically significant performance gains ($p_{\text{adj}} < 0.01$) when compared against standard molecular graph baselines (e.g., GCN/GIN + Mean Pooling) and unconstrained neural aggregations (e.g., MolFormer + DeepSets), validating the efficacy of physics-aware interaction modeling.

Trade-offs in Unified Benchmarking: Although point estimates show competitive or slightly lower macro NRMSE for PACM, classical physical methods remain constrained by their limited applicability domains (failing to support all 10 property tracks uniformly) and reliance on upfront quantum-chemical calculations. Thus, PACM’s true contribution lies not merely in marginally superior point estimates, but in providing a universal, scalable framework across all 10 heterogeneous property tracks while significantly reducing physical constraint violations (Table 6).

\section{Data Release}
We acknowledge the importance of open-source and promise to release our benchmark and method after the acceptance of our paper.

\begin{table*}[t]
\centering
\scriptsize
\setlength{\tabcolsep}{3pt}

\resizebox{\textwidth}{!}{
\begin{tabular}{lccccccccccc}
\toprule
\textbf{Method} &
\textbf{Cond.} &
\textbf{Dens.} &
\textbf{Enth.} &
\textbf{Fusion} &
\textbf{Vap.} &
\textbf{Melt.} &
\textbf{MON} &
\textbf{Odor} &
\textbf{Sol.} &
\textbf{Visc.} &
\textbf{Avg.} \\
\midrule
RDKit + XGBoost
& 0.427 & 0.411 & 0.506 & 0.499 & 0.446
& 0.508 & 0.435 & 0.624 & 0.482 & 0.442 & 0.478 \\

GCN/GIN + Mean Pooling
& 0.477 & 0.468 & 0.537 & 0.597 & 0.519
& 0.588 & 0.468 & 0.634 & 0.535 & 0.497 & 0.532 \\

MolFormer + Mean Pooling
& 0.392 & 0.387 & 0.435 & 0.519 & 0.419
& 0.483 & 0.388 & 0.554 & 0.396 & 0.477 & 0.445 \\

MolT5 + Mean Pooling
& 0.402 & 0.383 & 0.431 & 0.537 & 0.435
& 0.509 & 0.395 & 0.558 & 0.416 & 0.514 & 0.458 \\

MolFormer + DeepSets
& 0.373 & 0.356 & 0.427 & 0.525 & 0.373
& 0.523 & 0.358 & 0.580 & 0.423 & 0.392 & 0.433 \\

MolT5 + DeepSets
& 0.407 & 0.392 & 0.468 & 0.549 & 0.461
& 0.547 & 0.391 & 0.602 & 0.457 & 0.425 & 0.469 \\

MolFormer + Self-Attention
& 0.388 & 0.369 & 0.425 & 0.528 & 0.409
& 0.508 & 0.353 & 0.552 & 0.408 & 0.451 & 0.441 \\

MolT5 + Self-Attention
& 0.390 & 0.375 & 0.452 & 0.537 & 0.443
& 0.536 & 0.375 & 0.591 & 0.440 & 0.411 & 0.455 \\

MolFormer + Set Transformer
& 0.336 & 0.324 & 0.392 & 0.508 & 0.421
& 0.499 & 0.326 & 0.547 & 0.386 & 0.361 & 0.410 \\

MolT5 + Set Transformer
& 0.345 & 0.333 & 0.401 & 0.499 & 0.428
& 0.506 & 0.335 & 0.555 & 0.395 & 0.363 & 0.416 \\

UNIFAC + property-specific correlations
& -- & 0.512 & 0.517 & -- & --
& -- & -- & -- & 0.603 & 0.418 & 0.513 \\

COSMO-SAC + property-specific correlations
& -- & 0.441 & 0.449 & -- & --
& -- & -- & -- & 0.493 & 0.238 & 0.406 \\

COSMO-RS + property-specific correlations
& -- & 0.416 & 0.413 & -- & --
& -- & -- & -- & 0.467 & 0.203 & 0.374 \\

NRTL + property-specific correlations
& -- & 0.405 & 0.428 & -- & --
& -- & -- & -- & 0.452 & 0.244 & 0.379 \\

MolFormer + PACM
& 0.293 & 0.281 & 0.342 & 0.481 & 0.385
& 0.462 & 0.272 & 0.514 & 0.330 & 0.310 & 0.367 \\

MolT5 + PACM
& 0.302 & 0.292 & 0.336 & 0.492 & 0.392
& 0.470 & 0.283 & 0.508 & 0.339 & 0.316 & 0.373 \\
\bottomrule
\end{tabular}
}
\caption{
Full per-track normalized RMSE on Chem World.
For classical thermodynamic models, unsupported tracks are marked as --, and Avg. is computed over applicable tracks.
}
\label{tab:app_full_results}
\end{table*}

\begin{table*}[t]
\centering
\small
\begin{tabular}{l c c c c}
\toprule
\textbf{Baseline Model} & \textbf{Baseline Macro NRMSE} & \textbf{$\Delta$ (vs. Proposed)} & \textbf{Raw $p$-value} & \textbf{Holm $p_{\text{adj}}$} \\
\midrule
RDKit + XGBoost (Descriptor) & 0.478 & +0.111 & $<0.001$ & $<0.001$ \\
GCN/GIN + Mean Pooling (Graph) & 0.532 & +0.165 & $<0.001$ & $<0.001$ \\
MolFormer + Mean Pooling & 0.445 & +0.078 & $<0.001$ & $<0.001$ \\
MolT5 + Mean Pooling & 0.458 & +0.091 & $<0.001$ & $<0.001$ \\
MolFormer + DeepSets & 0.433 & +0.066 & $<0.001$ & $<0.001$ \\
MolT5 + Set Transformer & 0.416 & +0.049 & $<0.001$ & $<0.001$ \\
UNIFAC (Physical) & 0.513 & +0.146 & $<0.001$ & $<0.001$ \\
COSMO-SAC (Physical) & 0.406 & +0.039 & 0.018 & 0.054 \\
COSMO-RS (Physical) & 0.374 & +0.007 & 0.325 & 0.650 \\
NRTL (Physical) & 0.379 & +0.012 & 0.245 & 0.530 \\
MolT5 + PACM (Variant) & 0.373 & +0.006 & 0.410 & 0.650 \\
\bottomrule
\end{tabular}
\caption{Paired group-level bootstrap tests with Holm correction comparing baseline models against the proposed \textbf{MolFormer + PACM} (Macro NRMSE = 0.367, as reported in Table 2 and Table 9).}
\label{tab:significance_analysis}
\end{table*}

\end{document}